\documentclass[preprint,12pt,numbers,sort&compress]{elsarticle}

\usepackage{titlesec}

\titleformat{\paragraph}
  {\bfseries\itshape} 
  {} 
  {0pt} 
  {} 

\titleformat{\subparagraph}
  {\itshape} 
  {} 
  {0pt} 
  {} 
\usepackage{amssymb}
\usepackage{amsmath}
\usepackage{amsfonts}
\usepackage{array}
\usepackage{amssymb}
\usepackage{longtable}
\usepackage{xspace}
\usepackage{booktabs}
\usepackage{makecell}
\usepackage
{xcolor} 
\usepackage{graphicx}
\usepackage[margin=1in]{geometry} 
\usepackage[font=small]{caption}
\usepackage{booktabs}
\usepackage[unicode]{hyperref}
\usepackage{scrextend}
\usepackage{enumerate}
\usepackage{url}
\usepackage{tikz}
\usepackage{float}
\usepackage{ragged2e} 
\usepackage{afterpage}
\usepackage{multirow}
\usepackage{tocloft,calc}
\usepackage{listings}
\usepackage{makecell}
\usepackage{algorithm}
\usepackage{bookmark}
\usepackage{pifont}
\usepackage{perpage}
\usepackage{lmodern}
\usepackage{mathtools}
\usepackage{subcaption}
\usepackage{multicol}
\usepackage{wrapfig}
\usepackage{afterpage}
\usepackage{longtable}

\begin{document}

\begin{frontmatter}

\title{Reperio-rPPG: Relational Temporal Graph Neural Networks for Periodicity Learning in Remote Physiological Measurement}


\author[1]{Ba-Thinh Nguyen}
\author[1]{Thach-Ha Ngoc Pham}
\author[1]{Hoang-Long Duc Nguyen}
\author[1]{Thi-Duyen Ngo}
\author[1]{Thanh-Ha Le\corref{cor1}}

    \affiliation[1]{organization={Faculty of Information Technology, VNU University of Engineering and Technology},
                state={Ha Noi},
                country={Viet Nam}}

    \cortext[cor1]{Corresponding author:  ltha@vnu.edu.vn}

\begin{abstract}
Remote photoplethysmography (rPPG) is an emerging contactless physiological sensing technique that leverages subtle color variations in facial videos to estimate vital signs such as heart rate and respiratory rate. This non-invasive method has gained traction across diverse domains, including telemedicine, affective computing, driver fatigue detection, and health monitoring, owing to its scalability and convenience. Despite significant progress in remote physiological signal measurement, a crucial characteristic—the intrinsic periodicity—has often been underexplored or insufficiently modeled in previous approaches, limiting their ability to capture fine-grained temporal dynamics under real-world conditions. To bridge this gap, we propose Reperio-rPPG, a novel framework that strategically integrates Relational Convolutional Networks with a Graph Transformer to effectively capture the periodic structure inherent in physiological signals. Additionally, recognizing the limited diversity of existing rPPG datasets, we further introduce a tailored CutMix augmentation to enhance the model’s generalizability. Extensive experiments conducted on three widely-used benchmark datasets—PURE, UBFC-rPPG, and MMPD—demonstrate that Reperio-rPPG not only achieves state-of-the-art performance but also exhibits remarkable robustness under various motion (e.g., stationary, rotation, talking, walking) and illumination conditions (e.g., nature, low LED, high LED). The code is publicly available at https://github.com/deconasser/Reperio-rPPG.
\end{abstract}


\end{frontmatter}



\section{Introduction}
\label{sec:introduction}
Heart rate (HR) is a vital physiological signal that can indicate many important states of the body in health assessment. Traditional HR monitoring techniques, such as electrocardiogram (ECG) and photoplethysmography (PPG) \cite{Shaffer2017HRVreview, Malik1996HRVguidelines}, often use contact-based electrodes or wearable devices attached to the skin to detect this signal. These methods offer high accuracy but can lead to inconvenience during prolonged use or for individuals with sensitive, burned, or allergic skin conditions. Recently, non-contact HR measurement via rPPG has gained attention \cite{rppg_principle_1, rppg_principle_2, rppg_principle_3}, using facial videos captured through RGB cameras to extract subtle skin color changes by blood volume fluctuations. This approach overcomes the limitations of contact-based methods and enables many applications in telemedicine, driver monitoring, or fitness \cite{Verkruysse2008RemotePPG, McDuff2014TelemedicineRPPG, Al-Naji2022TelemedicineSurvey}. Despite their potential, non-contact methods remain sensitive to factors such as lighting conditions, skin tone variations, and subject movement. Although advances in signal processing and computer vision have mitigated some of these issues, achieving stable and accurate performance across real-world environments remains an open challenge, requiring further research and innovation.

Initial methods for non-contact HR estimation relied on classical signal processing techniques, such as blind source separation \cite{BSS_PCA_1, BSS_PCA_2, BSS_ICA_1} and color space transformation models \cite{POS, CHROM}, which analyze subtle color variations in facial videos to extract physiological signals. Later, with the rapid advancement of deep learning, particularly in computer vision, CNN-based approaches \cite{DeepPhys, TS-CAN} have gained significant attention for their ability to learn representative spatial and spatiotemporal features directly from raw video data. More recently, Transformer-based architectures \cite{PhysFormer, DINO-rPPG} have been introduced to this domain, leveraging self-attention mechanisms to further enhance the modeling of temporal dynamics in remote physiological signal estimation.

Most recent rPPG models are capable of extracting meaningful spatiotemporal features to estimate HR signals. However, many of these approaches treat the Blood Volume Pulse (BVP) as a generic temporal signal, without fully leveraging its fundamental characteristic—the quasi-periodic nature \cite{Period_1, Period_2}. Since heartbeats occur cyclically, the BVP exhibits structured patterns both within each cycle (intra-cycle) and across successive cycles (inter-cycle), which are crucial for accurate estimation. To approximate this behavior, prior studies have introduced techniques such as shift-based method \cite{TCS, TPS}, attention-based architectures \cite{TS-CAN, LSTM-SQA}, and frequency-aware designs \cite{DINO-rPPG, Contrast-Phys}. Despite these advances, existing approaches still struggle to capture the inherent regularity and phase consistency of physiological signals, underutilizing cross-cycle patterns that are key to modeling true periodicity.

To address these gaps, we take a graph-based view of temporal modeling. We treat frames as nodes and connect them with edges that encode how they interact over time and across phases. This lets the model adapt relations rather than rely on fixed shifts, capture both local and global structure, and aggregate long-range, phase-consistent context. Graphs are well-suited for learning hidden relations in temporal data, making them particularly appropriate for rPPG tasks. Yet, explicit graph formulations for periodicity in rPPG remain underexplored. To this end, we introduce Reperio-rPPG, a framework that can directly model the periodicity of HR. In summary, our method offers the following key contributions:

1. We propose a novel architecture combining a Swin Transformer for robust spatial feature extraction with a Relational Graph Convolutional Network (R-GCN) and a Graph Transformer, specifically designed to capture the periodic characteristics and temporal dependencies of rPPG signals.

2. We introduce an effective data augmentation strategy combining Temporal CutMix (TCM), Normalized Difference Frames (NDF) and Multi-Scale Plane-Orthogonal-to-Skin (MPOS), significantly improving the robustness and generalization capability of our rPPG model.

3. We conduct extensive experiments on three benchmark datasets—PURE, UBFC-rPPG, and MMPD—demonstrating that our proposed method consistently and substantially outperforms state-of-the-art approaches under both controlled laboratory conditions and challenging real-world scenarios characterized by diverse illumination variations, subject motion, and complex backgrounds.

\section{Related Work}
\subsection{Remote physiological measurement}

Early developments in rPPG primarily relied on traditional signal processing techniques, which can be categorized into blind signal separation methods and skin reflectance-based models. The former \cite{BSS_PCA_1, BSS_PCA_2, BSS_ICA_1} decomposes facial video signals into underlying components to isolate the cardiac pulse, while the latter \cite{POS, CHROM} enhances blood flow signals by transforming the signal into specific color spaces derived from skin reflectance models. These conventional methods are computationally efficient and demonstrate strong performance under controlled conditions (e.g., minimal motion, consistent lighting). However, their reliance on handcrafted features limits their robustness, where factors such as motion artifacts and lighting variations significantly degrade performance, rendering them unsuitable for practical, real-world applications.

Recent advances in deep learning have significantly improved rPPG estimation by enabling data-driven feature extraction from facial videos. Early approaches primarily leveraged 2D CNNs to extract spatial features, with models like DeepPhys \cite{DeepPhys} and TS-CAN \cite{TS-CAN} incorporating simple temporal mechanisms to enhance performance under constrained conditions. To better capture temporal dynamics, subsequent works introduced 3D CNN-based frameworks, such as PhysNet \cite{PhysNet}, AutoHR \cite{AutoHr}, and rPPGNet \cite{rPPGNet}, which jointly model spatiotemporal features across video frames. These methods demonstrated improved robustness and accuracy, particularly in handling fine-grained motion and physiological variations. Transformer architectures have recently been employed to model long-range spatiotemporal dependencies. PhysFormer \cite{PhysFormer} was among the first to introduce a video transformer for rPPG by leveraging temporal difference attention to improve signal modeling, achieving performance comparable to or exceeding CNN-based baselines. Subsequent models such as RADIANT \cite{RADIANT}, DINO-rPPG \cite{DINO-rPPG}, TransPPG \cite{TransPPG}, and RhythmFormer \cite{RhythmFormer} have demonstrated promising results across a range of conditions, highlighting the efficiency and adaptability of Transformer-based designs in this domain.

Beyond these architectures, several studies have introduced innovative designs for rPPG. PhysMamba \cite{PhysMamba} utilizes a state-space duality framework to better capture long-range temporal dependencies, while GraphPhys \cite{GraphPhys} and STGNet \cite{STGNet} adopt graph neural networks to ensure spatially consistent and noise-suppressed rPPG signals across facial ROIs. Both models \cite{GraphPhys} and \cite{STGNet} leverage the relational structure between neighboring regions to enforce coherence, with STGNet further extending this idea to the spatiotemporal domain to enhance stability across time.

\subsection{Temporal and periodic representation learning}

Accurate HR estimation in rPPG relies on modeling temporal dependencies in facial video streams. Early studies explored efficient temporal interaction modules, such as the Temporal Shift Module (TSM) and Temporal Channel Shift (TCS), originally proposed for video recognition tasks. These modules enhance temporal representation by mixing information across neighboring frames with minimal computational overhead. More recently, the Temporal Patch Shift (TPS) mechanism has been introduced to enable patch-level feature propagation across frames, facilitating spatiotemporal interaction with low latency and improved robustness to motion noise.

Beyond these shift-based designs, several works have focused on capturing richer temporal dynamics. Architectures inspired by SlowFast networks jointly model rapid and gradual temporal variations, while recurrent frameworks based on LSTM layers exploit sequential dependencies to recover consistent pulse patterns over time. Together, these methods have expanded the temporal modeling spectrum in rPPG, though their ability to generalize under real-world motion and illumination fluctuations remains limited.

In parallel, efforts toward learning periodic representations—core to the quasi-cyclic nature of cardiac signals—remain relatively underexplored. Multi-scale temporal modeling and frequency-domain regularization have been proposed to better represent periodicity, yet current methods still struggle to maintain stable phase alignment and consistent rhythm under challenging conditions. This underscores a key research direction: developing models that can jointly capture temporal dependencies and intrinsic physiological periodicity in complex, noisy environments.

\subsection{Graph Neural Networks}

Graph Neural Networks (GNNs) provide a powerful paradigm for learning from structured and relational data by representing components as nodes and their interactions as edges. Through iterative message passing and neighborhood aggregation \cite{kipf2017semi, velivckovic2018graph, wu2021comprehensive}, GNNs effectively capture both local and global dependencies within complex systems. This property has made them highly effective in domains characterized by structured interrelations, such as social networks \cite{hamilton2017inductive}, molecular and biological graphs \cite{gilmer2017neural}, spatiotemporal dynamics in traffic forecasting \cite{yu2018spatio} and human motion analysis \cite{yan2018stgcn}.

From a theoretical perspective, rPPG can naturally be formulated as a temporal graph learning problem. Each frame corresponds to a node carrying temporal features that reflect BVP variations, while temporal edges represent relations between frames. This structure aligns well with the GNN framework’s capacity to learn hidden relational dependencies and aggregate physiological cues across multiple cardiac cycles.

Despite this suitability, the use of GNNs in rPPG remains relatively underexplored. To date, only a few works, such as GraphPhys \cite{GraphPhys} and STGNet \cite{STGNet}, have leveraged graph-based architectures to reinforce an assumption that all facial ROIs encode the same underlying rPPG signal. However, neither approach utilizes the graph to explicitly represent or learn the periodic nature inherent to physiological signals.

Motivated by these insights, our work extends this line of research by designing a graph-based framework that not only models frame-to-frame relationships but also incorporates periodicity-aware temporal connectivity, enabling the network to capture both short-term dynamics and the cyclic patterns inherent in physiological signals.

\section{Method}
\begin{figure}[hbt] 
    \centering
    \includegraphics[width=\linewidth]{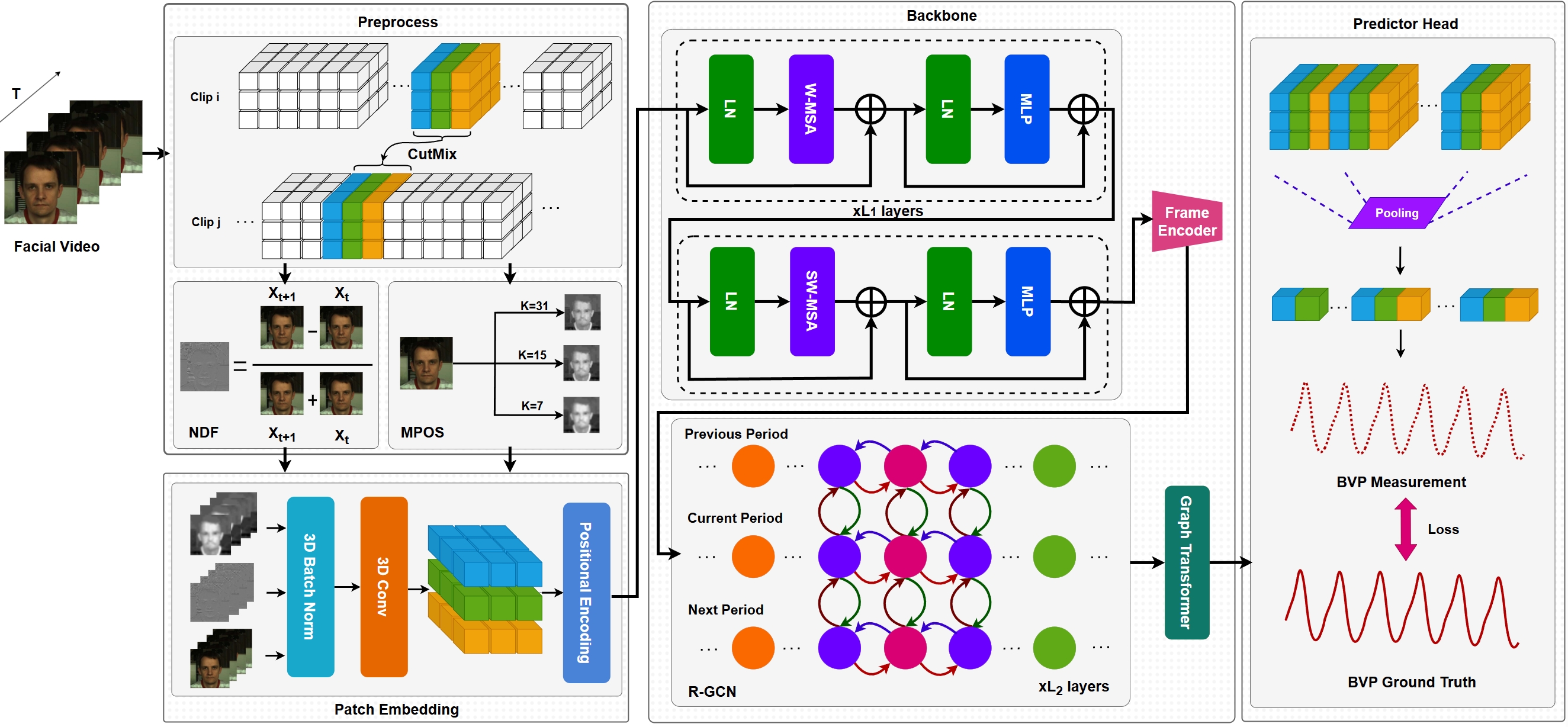} 

    \caption{Reperio-rPPG Architecture.}

    \label{fig:Reperio-rPPG Architecture} 
\end{figure}

As illustrated in Fig.~\ref{fig:Reperio-rPPG Architecture}, the proposed Reperio-rPPG framework consists of four key components. First, the Input Processing module pre-processes raw video frames to prepare them for further analysis, as detailed in Section~\ref{Input Processing}. Next, the Spatial Modeling block captures spatial features critical for physiological signal extraction (Section~\ref{Spatial Modeling}). Then, the Temporal Modeling module learns temporal dependencies across frame sequences (Section~\ref{temporal modeling}). Finally, the Predictor Head generates the rPPG signal from the learned spatiotemporal representations (Section~\ref{predictor head}).
\subsection{Input Processing}
\label{Input Processing}
Given a raw facial video, it is first divided into a sequence of short clips. Each clip and its label are represented as
\[
(\mathcal{C}, \mathcal{L}) =
\left(
  \{f_t\}_{t=1}^{T},\ \{l_t\}_{t=1}^{T}
\right), \quad
f_t \in \mathbb{R}^{H \times W \times 3},\quad l_t \in \mathbb{R},
\]
where \( T \) denotes the number of frames per clip, \( f_t \) is an RGB image with resolution \( H \times W \), and \( l_t \) corresponds to the GT target of frame \( f_t \).
 To enhance generalization in Reperio-rPPG, we introduce a tailored TCM augmentation strategy, in which a continuous temporal segment from one clip is replaced by a corresponding segment from another clip within the same batch. From each augmented clip, we further extract two complementary representations: the NDF \cite{ChannelWise, LSTS} and the MPOS \cite{LSTS}. The motivations and design principles behind these methods are discussed in the following subsections.
\subsubsection{Temporal CutMix (TCM)}
\begin{figure}[hbt]
    \centering
    \includegraphics[width=0.6\linewidth]{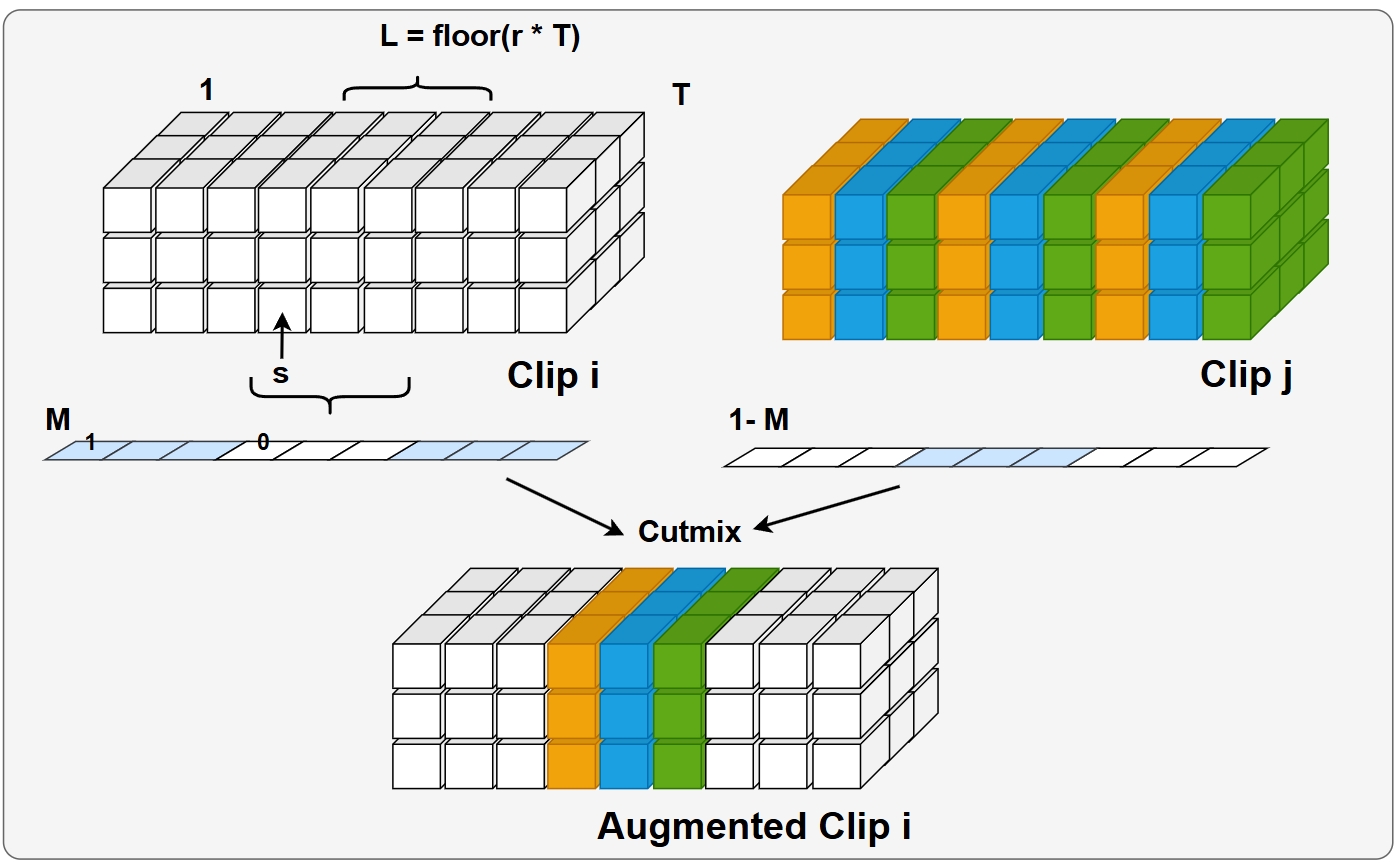}
    \caption{An illustrative example of Temporal CutMix.  The augmented Clip i is derived by replacing a temporal region from Clip i with that of Clip j (both from the same batch).}
    \label{fig:Cutmix}
\end{figure}
\paragraph{Formulation}
We apply TCM with probability \( p \) to a batch of \( B \) clips, where each clip is defined as \( \mathcal{C}^{i} = \{f^{i}_1, \dots, f^{i}_T\} \in \mathbb{R}^{T \times H \times W \times 3} \), with \( i \in \overline{1, B} \). When augmentation is triggered, a second clip \( \mathcal{C}^{j} \), where \( j \in \overline{1, B} \setminus \{i\} \), is randomly selected to serve as the replacement source (see Fig.~\ref{fig:Cutmix}). We then compute the number of frames \( L \in \mathbb{N} \) to be replaced in \( \mathcal{C}^i \) as:
\begin{equation}
L = \left\lfloor r \cdot T \right\rfloor, \quad r \sim \mathcal{U}(r_{\min}, r_{\max}),
\label{eq:L}
\end{equation}
where \( \mathcal{U} \) denotes the uniform distribution, \( \lfloor \cdot \rfloor \) is the floor operator, and \( r \) controls the ratio of frames to be replaced, with bounds applied to avoid excessively short or long segments.
To determine the temporal location of the replacement, we uniformly sample a starting index \( s \sim \mathcal{U}\{1, \dots, T - L + 1\} \). Based on this, we define a binary temporal mask \( M \in \{0,1\}^{T \times 1 \times 1 \times 1} \) as follows:
\begin{equation}
M_t =
\begin{cases}
0, & \text{if } s \leq t < s + L \\
1, & \text{otherwise}
\end{cases}
\label{eq:mask}
\end{equation}

The augmented clip and label are obtained by:
\begin{align}
\widetilde{\mathcal{C}}^i &= M \cdot \mathcal{C}^i + (1 - M) \cdot \mathcal{C}^j \label{eq:aug_clip} \\
\widetilde{\mathcal{L}}^i &= M \cdot \mathcal{L}^i + (1 - M) \cdot \mathcal{L}^j \label{eq:aug_label}
\end{align}

\begin{figure}[hbt]
    \centering
    \includegraphics[width=0.7\linewidth]{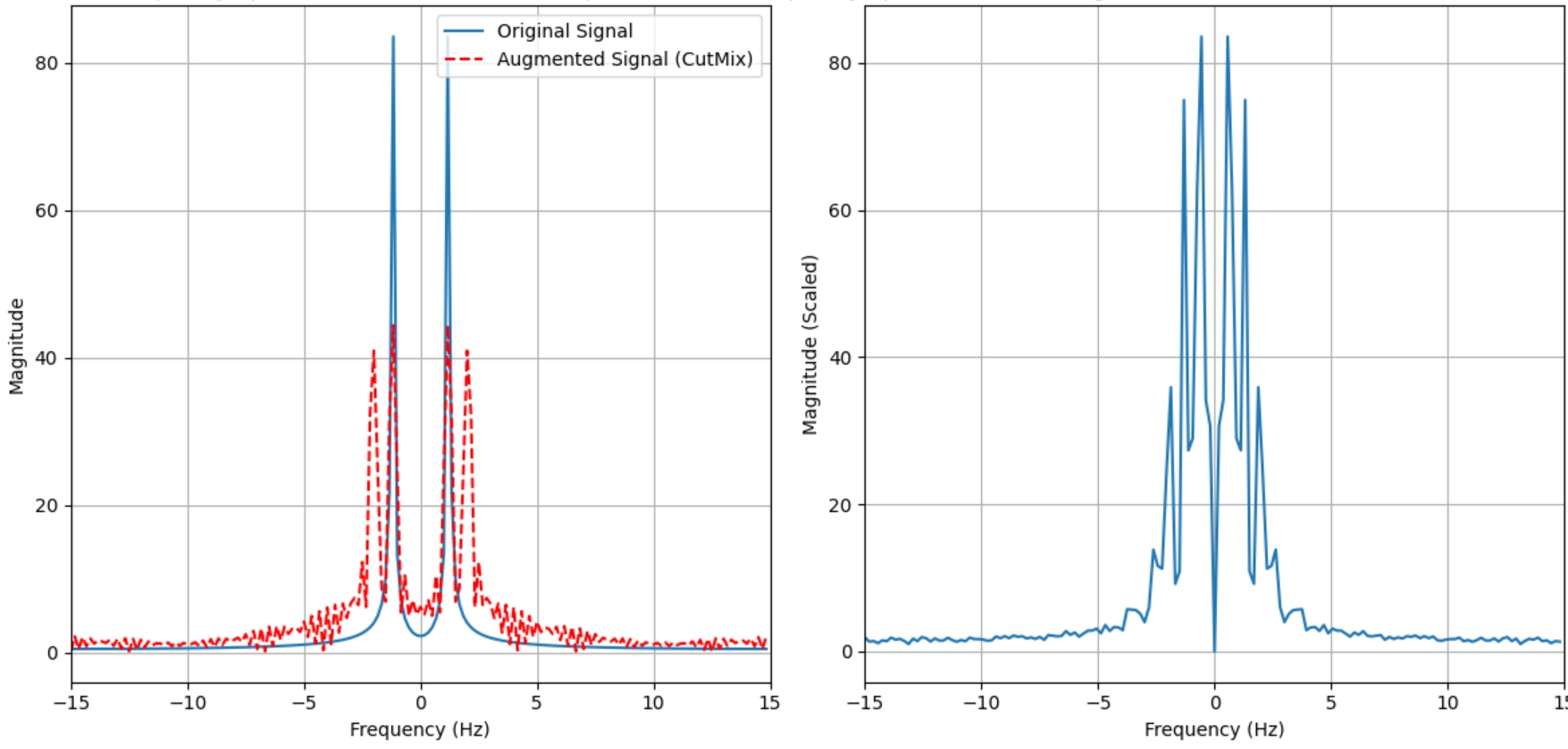}
    \caption{An illustrative example of frequency spectra before and after applying TCM [Left]. Real rPPG signal under fast translation from the PURE dataset [Right].}
    \label{fig:TCM}
\end{figure}

\paragraph{Theoretical Setup}
Previous studies have shown that CutMix acts as an implicit regularizer mixing inputs and soft labels to mitigate overfitting and enforce linear interpolations between classes \cite{provablebenefitcutoutcutmix, unifiedanalysismixedsample}. However, these studies focus on spatial-domain effects and overlook how temporal mixing shifts the input signal’s spectral content—a perspective that is particularly crucial for rPPG. In our study, we assume that the GT rPPG signal from clip \(\mathcal{C}^i\) can be modeled as an ideal sinusoid:
\begin{equation}
h^{i}(t) = A_i \sin(2\pi f_i t + \phi_i).
\label{eq:sine_signals}
\end{equation}
Before TCM, signal $h^{1}(t)$ has a Discrete-time Fourier transform (DTFT) characterized by impulses at $\omega_1 = \pm 2\pi f_1$ (see Fig.~\ref{fig:TCM}):
\begin{equation}
H^{1}(\omega) = \frac{A_1}{2i} \left[ e^{i\phi_1} \delta(\omega - \omega_1) - e^{-i\phi_1} \delta(\omega + \omega_1) \right].
\end{equation}
When applying TCM, a binary mask $M$ replaces a contiguous segment of length $L$ (starting from index $s$) in $h^{1}(t)$ with the corresponding segment from $h^{2}(t)$ as follows:
\begin{equation}
\widetilde{h}(t) = M \cdot h^{1}(t) + (1 - M) \cdot h^{2}(t).
\end{equation}
In the frequency domain, this mixing becomes a convolution between the DTFTs of the input signals and temporal mask:
\begin{equation}
\widetilde{H}(\omega) = \mathcal{F}\{M\} * H^{1}(\omega) + \mathcal{F}\{1 - M\} * H^{2}(\omega),
\label{eq:tcm_convolution}
\end{equation}
where $\mathcal{F}\{\cdot\}$ denotes the DTFT can be expressed as:
\begin{align}
\mathcal{F}\{M\}(\omega) &= \sum_{t=1}^{T} M(t)\,e^{-i\omega t} 
= \underbrace{\sum_{t=1}^{T} e^{-i\omega t}}_{U(\omega)} - \underbrace{\sum_{t=s}^{s+L-1} e^{-i\omega t}}_{W(\omega)}, \\
\mathcal{F}\{1-M\}(\omega) &= W(\omega).
\end{align}
The term $U(\omega)$ is a finite geometric series:
\begin{equation}
U(\omega) = e^{-i\omega}\,\frac{1-e^{-i\omega T}}{1-e^{-i\omega}}
= e^{-i\omega\frac{T+1}{2}} \frac{\sin(\tfrac{T\omega}{2})}{\sin(\tfrac{\omega}{2})},
\end{equation}
and similarly
\begin{equation}
W(\omega) = e^{-i\omega\frac{2s+L-1}{2}} \frac{\sin(\tfrac{L\omega}{2})}{\sin(\tfrac{\omega}{2})}.
\end{equation}
When $\omega$ approaches 0, applying L’Hôpital’s rule, we obtain:
\[
\lim_{\omega\to 0} U(\omega)=T, \quad \lim_{\omega\to 0} W(\omega)=L, \quad \lim_{\omega\to 0}\mathcal{F}\{M\}(\omega)=T-L.
\]
Finally, substituting back into the convolution expression~\ref{eq:tcm_convolution}, the mixed spectrum becomes:
\begin{align}
\widetilde{H}(\omega) 
&= \frac{A_1}{2i} \Big[ e^{i\phi_1}(U(\omega - \omega_1) - W(\omega - \omega_1))
 - e^{-i\phi_1}(U(\omega + \omega_1) - W(\omega + \omega_1)) \Big] \nonumber\\
&\quad + \frac{A_2}{2i} \Big[ e^{i\phi_2} W(\omega - \omega_2) - e^{-i\phi_2} W(\omega + \omega_2) \Big].
\end{align}

From the above expression, TCM introduces two key frequency-domain effects that highlight its utility for rPPG signals. First, the dominant spectral peaks at $\pm \omega_1$ are attenuated due to masking, with their magnitude scaled by $\mathcal{F}{\{M\}}(0) = U(0) - W(0) = T - L$, resulting in a reduction factor of $\frac{T - L}{T}$. In contrast, the component at $\omega_2$ does not appear as a sharp spike but rather as a lower-energy lobe, weighted proportionally by the mixing ratio $\frac{L}{T}$. Second, the convolution with DTFTs of rectangular windows generates sinc-shaped side lobes around both $\omega_1$ and $\omega_2$. These side lobes, appearing as "shoulders" around the main peaks, cause spectral leakage that mimics realistic frequency perturbations (e.g., motion or illumination variations). Thus, this frequency domain view explains why TCM is well-suited for rPPG; it intentionally introduces distortions aligned with common signal extraction challenges. As shown in Fig.~\ref{fig:TCM}, the augmented spectrum closely resembles that of a real rPPG signal under fast subject motion, thereby enhancing model robustness and generalization.
\subsubsection{Normalized Difference Frames (NDF)} Given a clip $\widetilde{\mathcal{C}^i} = \{f_1^i, f_2^i, \ldots, f_T^i\}$, we construct a normalized difference representation to emphasize subtle temporal variations in pixel intensity while suppressing illumination inconsistencies. Specifically, the NDF at time step $t$ is defined as:

\begin{equation}
\widetilde{\mathcal{C}}_{\text{D}, t}^i = \frac{f_{t+1}^i - f_t^i}{f_{t+1}^i + f_t^i}, \quad t < T, \quad \widetilde{\mathcal{C}}_{\text{D}, T}^i = 0.
\end{equation}

\subsubsection{Multi-Scale Plane-Orthogonal-to-Skin (MPOS)}
We employ the MPOS technique \cite{LSTS} to robustly extract rPPG signals. First, we normalize the augmented clip $\widetilde{\mathcal{C}^i}$ by its temporal mean and project it onto the POS space using the projection matrix from \cite{POS}:
\begin{equation}
P =
\begin{bmatrix}
0 & 1 & -1 \\
-2 & 1 & 1
\end{bmatrix},
\end{equation}
producing the transformed representation:
\begin{equation}
   \widetilde{\mathcal{C}^i}_{\text{POS}} = P^T \cdot \text{Norm}(\widetilde{\mathcal{C}^i}),
\end{equation}
where \( \text{Norm}(\cdot) \) denotes temporal average normalization.  Next, we extract multi-scale
spatial features via Gaussian blurring with kernel sizes \( k \in \{7, 15, 31\} \):
\begin{equation}\widetilde{\mathcal{C}}_{G,k}^i = G(\widetilde{\mathcal{C}}
^i_{\text{POS}}, k),
\end{equation}
where \( G(\cdot, k) \) is the Gaussian blur at  scale \( k \). For each scale, we compute the MPOS feature by combining the two channels of \( \widetilde{\mathcal{C}}_{G,k}^i \):
\begin{equation}
\widetilde{\mathcal{C}}_{\text{MPOS}, k}^i = \widetilde{\mathcal{C}}_{G,k}^i[\dots, 1] + \frac{\text{Std}(\widetilde{\mathcal{C}}_{G,k}^i[\dots, 1])}{\text{Std}(\widetilde{\mathcal{C}}_{G,k}^i[\dots, 2])} \widetilde{\mathcal{C}}_{G,k}^i[\dots, 2],
\end{equation}
where \( [\dots, i] \) represents  the \( i \)-th channel and \( \text{Std}(\cdot) \) is the temporal standard deviation. Finally, we concatenate these multi-scale features:
\begin{equation}
    \widetilde{\mathcal{C}}^i_{\text{MPOS}} = \text{Concat} \left( \widetilde{\mathcal{C}}_{\text{MPOS}, 7}^i, \widetilde{\mathcal{C}}_{\text{MPOS},15}^{i}, \widetilde{\mathcal{C}}_{\text{MPOS}, 31}^{i} \right).
\end{equation}
\subsubsection{Final Input Representation}
We construct the final feature map by concatenating the TCM, NDF, and MPOS features as follows:

\begin{equation}
\mathcal{C}_{aug} = \text{Concat}(\widetilde{\mathcal{C}}^i_{\text{TCM}}, \widetilde{\mathcal{C}}_{\text{D}}^i, \widetilde{\mathcal{C}}_{\text{MPOS}}^i).
\end{equation}

This map is partitioned into non-overlapping  \( 16 \times 16 \) spatial patches, yielding dimensions \( H' = H / 16 \) and \( W' = W / 16 \). A 3D convolutional layer subsequently embeds these patches into a  \( D \)-dimensional feature space, followed by layer normalization (\(\mathcal{LN}\)) \cite{LayerNorm} and positional encoding (\(\mathcal{PE}\)) \cite{AttentionIsAllYouNeed}:

\begin{equation}
\mathcal{C}_{\text{embed}} = \mathcal{LN}(\mathcal{PE}(\text{Conv3D}(\mathcal{C}_{aug}))).
\end{equation}

The resulting sequence $\mathcal{C}_{\text{embed}} \in \mathbb{R}^{B \times D\times T \times H' \times W'}$ serves as input to the spatial modeling module. 
\subsection{Spatial Modeling}
\label{Spatial Modeling}
 In this study, we adopt the Swin Transformer architecture, which utilizes window-based hierarchical self-attention to capture both local facial details and broader contextual information across facial regions. To enhance robustness against high-frequency noise, we employ a single-stage configuration with relatively large patch sizes. This design balances detail preservation with spatial smoothing, ensuring stable feature extraction under minor motion or illumination variations. The output is a sequence of spatial embeddings $\{f_1, f_2, \dots, f_T\}$, where each $f_i \in \mathbb{R}^{B \times D \times H' \times W'}$ represents the spatial features of frame $i$. These features are subsequently passed to the temporal modeling module to capture periodic dependencies over time.
\subsection{Temporal Modeling}
\begin{figure}[hbt]
    \centering
    \includegraphics[width=0.5\linewidth]{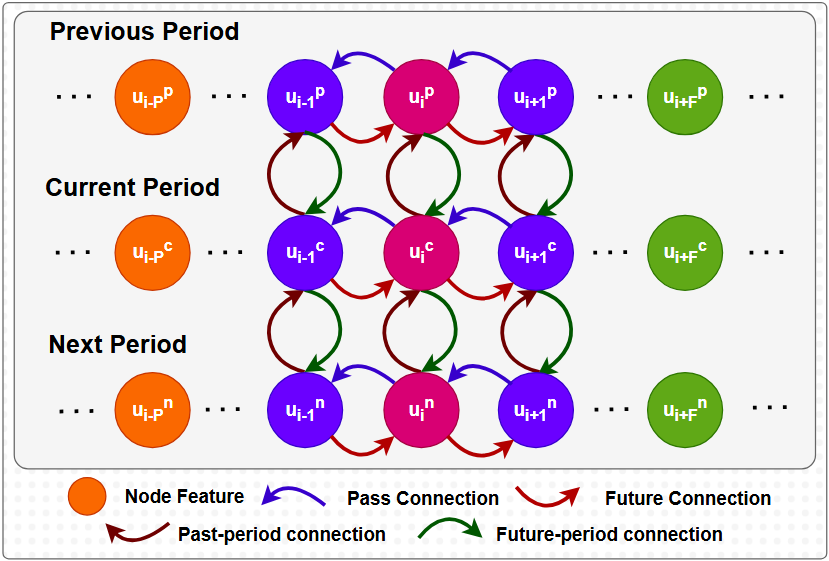}
    \caption{An illustrative example of a graph construction for the query node $u_i^c$ with a window size of $[\mathcal{P}, \mathcal{F}] = [1, 1]$.}
    \label{fig:RGCN}
\end{figure}
\label{temporal modeling}
In rPPG, consecutive frames often appear visually similar but reflect varying BVP values due to subtle physiological changes. The BVP signal is quasi-periodic with high temporal variability, making accurate waveform prediction challenging without broader contextual information. To achieve high correlation with such variable targets, each frame’s prediction must be guided by physiologically relevant context that helps the model disambiguate subtle fluctuations and rhythmic patterns. We address this by introducing a graph-based framework combining R-GCN and Graph Transformer, where each node dynamically aggregates information from meaningful neighbors within and across heartbeat cycles. This design enables each frame to refine its temporal representation while preserving phase consistency, leading to more robust and context-aware predictions. The novelty of our approach lies in this structured, period-aware relational modeling, which has not been explored by existing graph-based methods in rPPG.

\textbf{Definition:} The temporal graph \(\mathcal{G}(\mathcal{V}, \mathcal{E}, \mathcal{R})\) is defined based on the spatial modeling output, where:  
\begin{itemize}
    \item \(\mathcal{V}\): the set of node features.  
    \item \(\mathcal{E}\): the set of edges connecting nodes.  
    \item \(\mathcal{R}\): the types of relation between nodes.  
\end{itemize}

\textbf{Node:} The feature map derived from the Swin Transformer, which can be described as:
\begin{equation}
    \mathcal{C}_{spatial} = \{ f_1, f_2, \dots, f_T \}.
\end{equation}
Here, each feature $f_i \in \mathbb{R}^{B \times D \times H' \times W' }$ is then processed by a Frame Encoder ($FE$) to obtain its corresponding vector representation. Formally, this can be expressed as:
\begin{equation}
x_i = \operatorname{Pool}_{\mathrm{avg}}^{3\mathrm{D}}(f_i), \quad \text{for } i = 1, \dots, T.
\end{equation}
Consequently, the full set of nodes is defined as:
\begin{equation}
X = \{ x_1, x_2, \dots, x_T \}, \quad \text{where} \quad x_i \in \mathbb{R}^{B  \times d_{g}}.
\end{equation}

\textbf{Edges:} Each edge $(x_i, x_j, r_{ij}) \in$ \(\mathcal{E}\) represents an interaction between nodes $x_i$ and $x_j$ with a relation type $r_{ij} \in$ \(\mathcal{R}\). In this study, we propose to categorize relations into two main groups: intra-period \(\mathcal{R}_{intra}\) and inter-period relations \(\mathcal{R}_{inter}\) (see Fig.~\ref{fig:RGCN}).
\subsubsection{Intra-Period Relations}
$\mathcal{R}_{\text{intra}}$ captures local interactions within a single heartbeat period. It is further divided into:

\begin{itemize}
    \item Past Contextual Window (\(\mathcal{P}\)), which represents connections from previous nodes to the current node.
    \item Future Contextual Window (\(\mathcal{F}\)), which represents connections from the current node to next nodes.
\end{itemize}
For clarity, the intra-period relation structure can be defined as:
\begin{equation}
\mathcal{R}_{intra} =
\left\{
\begin{array}{ll}
\{ x_j \overset{\text{past}}{\leftarrow} x_i \mid \max(1, i - \mathcal{P}) \leq  j < i \} \\
\{ x_i 
\overset{\text{future}}{\rightarrow} x_j \mid i < j \leq \min(T, i + \mathcal{F}) \}
\end{array}
\right\}
\end{equation}
where $\quad i,j \in \overline{1, T}; \quad 
\overset{\text{past}}{\leftarrow}
 \text{ and } 
\overset{\text{future}}{\rightarrow}
$ indicate the past and future relation respectively.

This formulation enables each node to effectively integrate contextual information from its neighboring frames within a given heartbeat period.
\subsubsection{Inter-Period Relations}
\begin{figure}[hbt] 
    \centering    
    \includegraphics[width=0.4\linewidth]{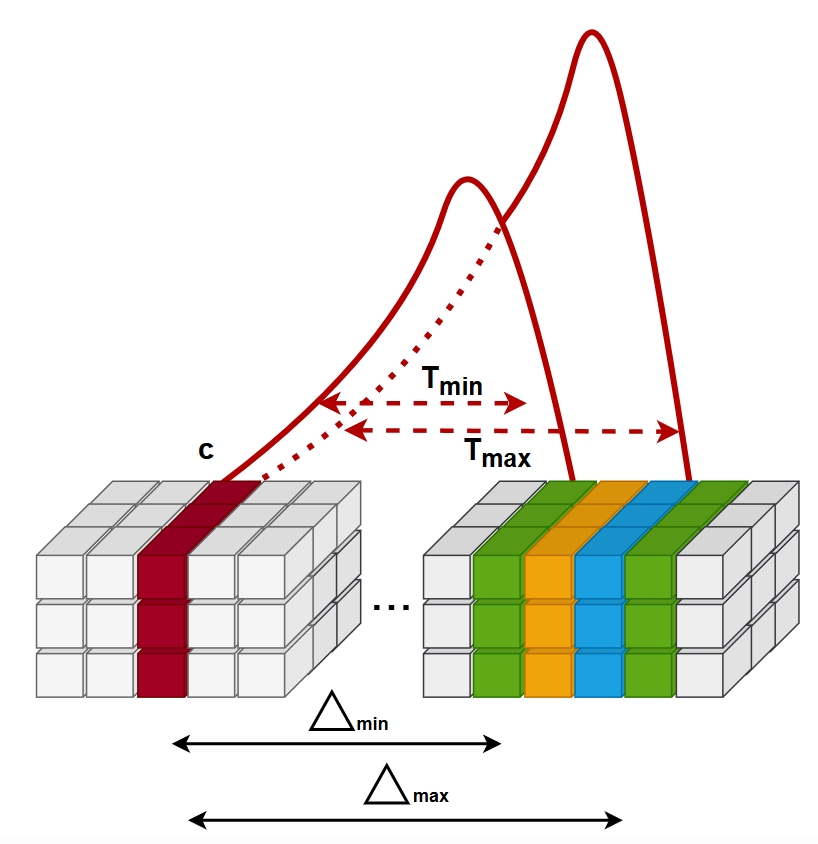} 
    \caption{An illustrative example of inter-period temporal dependencies across different heartbeat cycles. The typical duration of a human heartbeat lies within the physiological range \( T = [T_{\min}, T_{\max}] \), where \( T_{\min} \) and \( T_{\max} \) represent the minimum and maximum allowable durations of a normal cardiac cycle, respectively. Given the temporal resolution of the video (i.e., frames per second, fps), this time interval can be mapped to a specific range of video frames, denoted as \( \Delta_{\min} \) and \( \Delta_{\max} \).}
    \label{fig:Inter-Period Relations} 
\end{figure}
 \(\mathcal{R}_{inter}\) serves as the set of periodic interactions between nodes across different heartbeat cycles. The duration of a typical heartbeat falls within the physiological range \(T = [T_{\min}, T_{\max}]\) \cite{LSTS}, where \(T_{\min}\) and \(T_{\max}\) denote the minimum and maximum durations of a normal heartbeat, correspondingly. Given a reference time \textit{c}, we establish a temporal window based on this range to capture periodic dependencies effectively. As illustrated in Fig.~\ref{fig:Inter-Period Relations}, the lower bound \(\Delta_{\min}\) corresponds to \(T_{\min}\), while the upper bound \(\Delta_{\max}\) corresponds to \(T_{\max}\). Using these bounds, we formulate the inter-period relation \(\mathcal{R}_{inter}\) as follows:
\begin{equation}
\mathcal{R}_{inter}=
\left\{
\begin{aligned}
& \big\{ x_{j}^p\overset{\scriptscriptstyle{p}}{\leftarrow} x_{i}^c \mid \max(1, i - \Delta_{\max}) \leq j \leq i - \Delta_{\min} \big\} \\
& \big\{x_{i}^c\overset{\scriptscriptstyle{n}}{\rightarrow}x_{j}^n \mid i + \Delta_{\min} \leq j \leq \min(T, i + \Delta_{\max}) \big\}
\end{aligned}
\right\}
\end{equation}
where \(i,j \in \overline{1, T}\), $\quad 
\overset{\scriptscriptstyle{p}}{\leftarrow}
 \text{ and } 
\overset{\scriptscriptstyle{n}}{\rightarrow}
$ denote the past and next period relations, respectively; \( x_{i,j}^{\tau} \) with \( \tau \in \{p, c, n\} \), represents the node feature corresponds to the previous (\textit{p}), current (\textit{c}), and next (\textit{n}) period in that order. Each node is dynamically assigned a state depending on the reference time. Specifically, when evaluating a node \( x_i \) at time \( c \), it is considered as \( x_{i}^{c} \), while its neighboring nodes within the defined temporal window are assigned to either the past state (\( x_{j}^{p} \), if \( j < i \)) or the future state (\( x_{j}^{n} \), if \( j > i \)).

These relations allow the model to efficiently capture periodic dependencies, facilitating structured temporal reasoning across non-adjacent yet rhythmically aligned observations.

\textbf{Graph Learning:} In our framework, node representations are first updated by aggregating information from their neighboring nodes based on different relation types $r \in \mathcal{R}$. For each relation, a transformation function \( f(\mathbf{H}, \mathbf{W}_r) \) is applied, where \( \mathbf{W}_r\) is the learnable weight matrix specific to relation \textit{r}. The final node representation is obtained by integrating contributions from all relation types. Specifically, the feature update is defined as:
\begin{equation}
    g_i = \sum_{r \in \mathcal{R}} \sum_{j \in \mathcal{N}_r(i)} \frac{1}{|\mathcal{N}_r(i)|} \mathbf{W}_r \cdot x_j + \mathbf{W}_0 \cdot x_i,
\end{equation}
where $\mathcal{N}_r(i)$ denotes the set of neighboring nodes under relation $r$; \( \mathbf{W}_0\), \( \mathbf{W}_r\) $\in \mathbb{R}^{d_{h1} \times d_g}$ are learnable parameters (with $h_1$ representing the hidden layer dimension used by the R-GCN).

To further enhance the representational capacity of these node features, we incorporate a Graph Transformer module \cite{GraphTransformer, CORECT}. This module leverages a multi-head self-attention mechanism combined with feed-forward neural networks, allowing it to effectively capture both local and global structural dependencies. Given the representation $g_i$ produced by the R-GCN, the refined node features are computed as follows:

\begin{equation}
    o_i = \Bigg\|_{c=1}^{C} \left( \mathbf{W}_1 g_i + \sum_{j \in \mathcal{N}(i)} \alpha_{i,j} \mathbf{W}_2 g_j\right),
\end{equation}
where $\mathbf{W}_1^{c}, \mathbf{W}_2^{c} \in \mathbb{R}^{d_{h2} \times d_{h1}}$ are learnable matrices with $h_2$ denoting the hidden layer dimension of the Graph Transformer, $\mathcal{N}(i)$ is the set of nodes connected to node $i$, and $\Vert$ denotes the concatenation across $C$ attention heads. The attention coefficients $\alpha_{i,j}$ are computed using a scaled dot-product attention mechanism:
\begin{equation}
    \alpha_{i,j} = softmax \left( \frac{(\mathbf{W}_3 g_i)^\top (\mathbf{W}_4 g_j)}{\sqrt{d}} \right),
\end{equation}
where $\mathbf{W}_3^{c}, \mathbf{W}_4^{c} \in \mathbb{R}^{d_{h\alpha} \times d_{h1}}$ are learnable parameters.

After propagating and aggregating information throughout the graph, the final node representations are given by:

\begin{equation}
    G = \{ o_1, o_2, \dots, o_T\}, \quad \text{where} \quad o_i \in \mathbb{R}^{d_{h2} \times C}.
\end{equation}

This design effectively integrates local relational interactions with global temporal dependencies, ensuring the preservation of fine-grained and periodic characteristics that are crucial for robust rPPG analysis.

\subsection{Predictor Head}
\label{predictor head}
The predicted BVP signal, denoted by $\hat{\mathcal{L}}$, is obtained by applying a pooling operation to the node representations. Specifically, each node representation $o_i$ (for $i = 1, \dots, T$) is processed through a linear layer that outputs a single value. The resulting outputs are then concatenated to form $\hat{\mathcal{L}}$:

\begin{equation}
\hat{\mathcal{L}} = \big\|_{i=1}^{T} \text{Linear}(o_i, 1),
\end{equation}
where $\|$ denotes the concatenation operator.

To measure the discrepancy between $\hat{\mathcal{L}}$ and the GT BVP signal $\mathcal{L}$, we adopt the Negative Pearson Correlation Coefficient as the loss function following previous studies~\cite{LSTS, Physformer++}. It is defined as:
\begin{equation}
Loss = 1 - \frac{\text{cov}(\mathcal{L}, \hat{\mathcal{L}})}{\sqrt{\text{cov}(\mathcal{L}, \mathcal{L})} \cdot \sqrt{\text{cov}(\hat{\mathcal{L}}, \hat{\mathcal{L}})}},
\end{equation}
where $\text{cov}(\cdot)$ represents the covariance function.
\section{Experiments}
\subsection{Datasets}
We evaluate our methods using three widely-used datasets: PURE \cite{PURE}, UBFC-rPPG \cite{UBFC}, and MMPD \cite{MMPD}.

\textbf{PURE}: The PURE dataset \cite{PURE} is a high-quality benchmark for rPPG featuring 10 subjects (8 males, 2 females) performing six controlled head motions (sitting still, talking, slow head movement, quick head movement, small head rotation and medium head rotation). These varied activities simulate real-world scenarios, allowing researchers to assess how motion affects rPPG signal accuracy and the performance of HR estimation algorithms in practical settings. All videos were recorded at a resolution of 640×480 pixels at 30 Hz using lossless PNG encoding, and they were synchronized with GT measurements from a CMS50E pulse oximeter at 60 Hz. Despite its relatively small scale (60 videos), the PURE dataset remains widely adopted in rPPG research due to its controlled conditions and precision. It serves as a benchmark for evaluating motion robustness in HR estimation algorithms, particularly under varying head movements and facial motions.

\textbf{UBFC-rPPG}: The UBFC-rPPG dataset \cite{UBFC} consists of 42 videos recorded indoors, where subjects engaged in a time-sensitive mathematical game designed to induce natural HR variations while simulating typical human-computer interactions. Video data was captured using a webcam at a frame rate of 30 fps with a resolution of 640×480 pixels, while GT PPG waveforms and HR values were obtained with a CMS50E transmissive pulse oximeter. The dataset features natural variations in ambient lighting conditions, combining sunlight and artificial indoor illumination, which makes it more challenging than the PURE dataset.
\begin{figure}[H] 
    \centering
    \includegraphics[width=\linewidth]{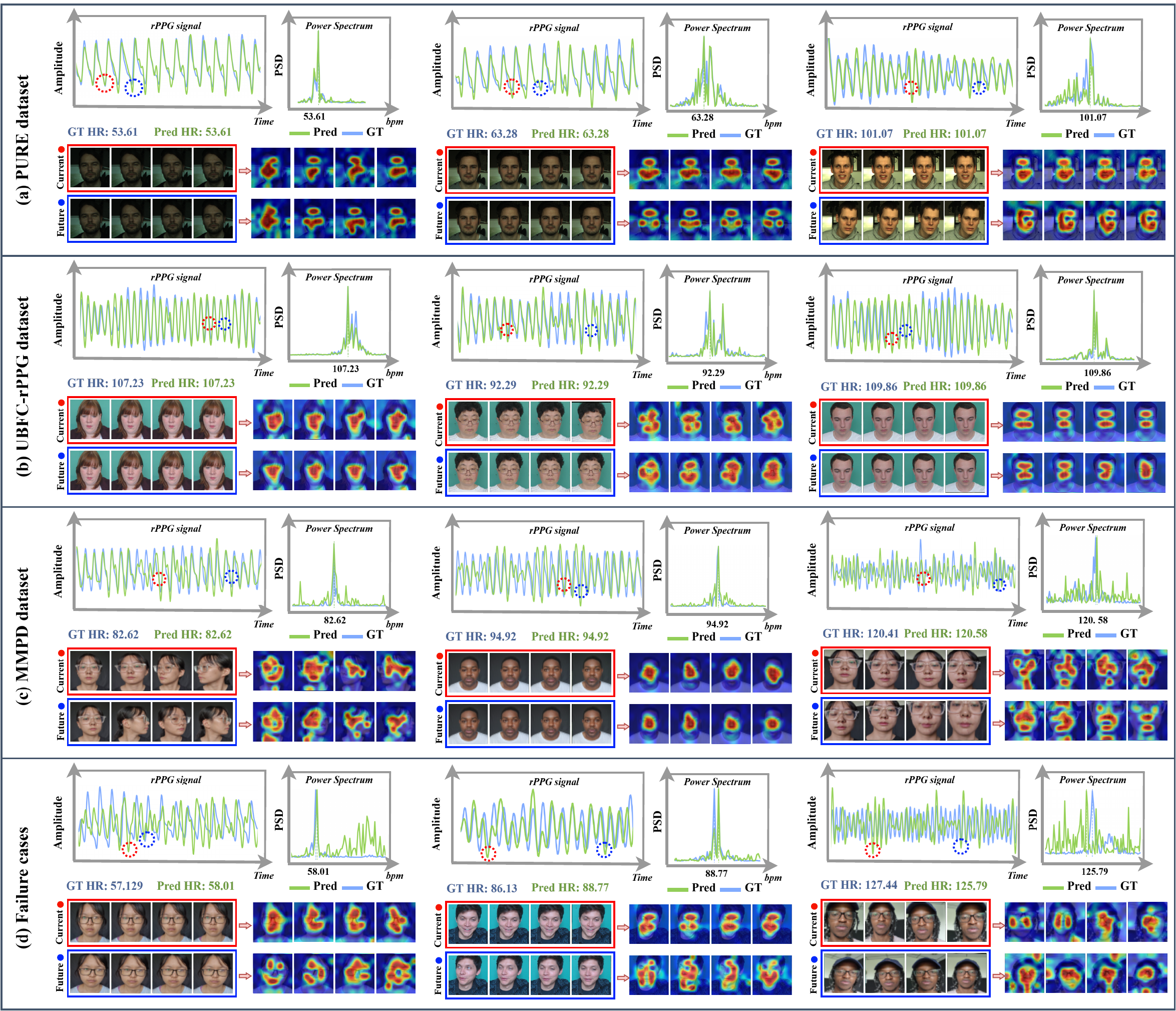}
    \caption{Visualization of HR prediction results on the (a) PURE, (b) UBFC-rPPG, and (c) MMPD datasets, together with several failure and challenging cases (d). The visualized samples represent a wide variety of subjects that differ in gender, age, skin tone, and illumination. They also cover multiple motion conditions, including steady posture, head movement, talking, and walking. The HR values observed in these samples range from low rates below 60 bpm to high rates exceeding 120 bpm, as well as normal physiological levels, ensuring a comprehensive evaluation of the model’s performance across diverse physical states. In Figures~\ref{fig:heatmap} (a)–(c), the predicted rPPG signals exhibit close temporal correspondence with the GT signals, particularly for HR values computed from PSD analysis. Representative frames sampled at the troughs of rPPG signals and their corresponding feature maps reveal spatial consistency and strong visual activation at the trough locations. Figures~\ref{fig:heatmap} (d) illustrate three failure examples drawn from UBFC-rPPG and MMPD, which emphasize the model’s limitations when confronted with skin tone,  extreme motion or facial expression variations. Overall, the visualization demonstrates the robustness and interpretability of Reperio-rPPG across diverse and realistic scenarios.}
    \label{fig:heatmap} 
\end{figure}

\textbf{MMPD}: The Multi-domain Mobile Video Physiology Dataset (MMPD) \cite{MMPD}, comprising approximately 370 GB of video data at a resolution of 320×240, serves as a comprehensive benchmark for rPPG research. It includes about 11 hours of video recordings collected from 33 subjects using mobile phone cameras. The dataset was specifically designed to address the limitations in existing rPPG datasets by systematically varying three key domains that influence algorithm performance: skin tone (Fitzpatrick types III–VI), lighting conditions (LED-high intensity, LED-low intensity, incandescent, natural lighting), and activity types (stationary, head rotation, talking, walking). These variations make MMPD one of the most diverse and challenging benchmarks available for real-world rPPG evaluation.

\subsection{Implementation Details}
\label{Implementation Details}
In this study, we utilize RetinaFace \cite{Retina} for facial detection. To ensure sufficient coverage of the facial region, we expand the detected bounding box by a factor of 1.5 before cropping and resizing it to a standardized resolution of 128 × 128 pixels. Face image frames are randomly flipped horizontally. A sliding window approach is then applied to generate 180-frame clips with a step size of 90 frames. For model configuration, we set the patch embedding dimension to $D = 192$; $p = 0.4$, $[r_{min}, r_{max}]=[0.25, 0.5]$ in TCM. The Swin Transformer backbone consists of $L_1 = 12$ layers with 8 attention heads. The Graph Transformer module is implemented with a single layer $L_2 = 1$, $d_g = 128$, $h_1 = 100$, $h_2 = 100$, along with 7 attention heads. The past and future window parameters in R-GCN $[\mathcal{P}, \mathcal{F}]$ are set to $[1, 1]$ for the PURE and UBFC-rPPG, and $[5, 5]$ for MMPD dataset. The periodicity window is defined as $[\Delta_{\text{min}}, \Delta_{\text{max}}] = [15, 25]$, which corresponds to a temporal frequency range of $[T_{\text{min}}, T_{\text{max}}] = [72, 120]$ bpm, as suggested in \cite{LSTS}.

All models are trained from scratch on an NVIDIA GeForce RTX 3090 GPU using PyTorch 2.0.0 with CUDA 11.8. We employ the Adam optimizer with a batch size of 4, an initial learning rate of \( 1 \times 10^{-4} \), and a weight decay of 0.01. The learning rate is dynamically adjusted using a OneCycle scheduler with a cosine annealing strategy. Training is conducted for 10 epochs on the MMPD dataset and for 30 epochs on both the PURE and UBFC-rPPG datasets.

\subsection{Metric and Evaluation}

To evaluate the performance of our proposed methods, we employ four widely used metrics: Mean Absolute Error (MAE), Root Mean Square Error (RMSE), Mean Absolute Percentage Error (MAPE), and Pearson’s Correlation Coefficient (r). The corresponding equations are defined as follows:

\begin{equation}
    \text{MAE} = \frac{1}{N} \sum_{i=1}^{N} |y_i - \hat{y}_i|
\end{equation}

\begin{equation}
    \text{RMSE} = \sqrt{\frac{1}{N} \sum_{i=1}^{N} (y_i - \hat{y}_i)^2}
\end{equation}

\begin{equation}
    \text{MAPE} = \frac{100\%}{N} \sum_{i=1}^{N} \left| \frac{y_i - \hat{y}_i}{y_i} \right|
\end{equation}

\begin{equation}
    r = \frac{\sum_{i=1}^{N} (y_i - \bar{y})(\hat{y}_i - \bar{\hat{y}})}{\sqrt{\sum_{i=1}^{N} (y_i - \bar{y})^2} \sqrt{\sum_{i=1}^{N} (\hat{y}_i - \bar{\hat{y}})^2}}
\end{equation}

where \( y_i \) and \( \hat{y}_i \) represent the GT and predicted values, in the same order; \( \bar{y} \) and \( \bar{\hat{y}} \) denote their corresponding mean values; and \( N \) is the total number of samples.

For Heart Rate Variability (HRV) evaluation, we consider three widely used metrics \cite{HRV_Measurement}. The Interbeat Interval (IBI) represents the time between successive heartbeats and is the fundamental signal from which HRV is derived. Standard Deviation of NN intervals (SDNN) quantifies the overall variability of heartbeats and reflects the combined influences of both sympathetic and parasympathetic activity. SD1/SD2 \cite{SD1SD2} is a non-linear metric, where SD1 reflects short-term variability and SD2 reflects both short and long-term components; their ratio offers insight into autonomic balance. Together, these metrics capture diverse aspects of cardiac rhythm dynamics in both time and non-linear domains.
\subsection{Main Comparison}
To assess the effectiveness of Reperio-rPPG, we compare it with existing methods through both intra-dataset and cross-dataset testing. The experiments are conducted on the PURE \cite{PURE}, UBFC-rPPG  \cite{UBFC}, and MMPD  \cite{MMPD} datasets. Details of the experimental setup are presented below.
\subsubsection{Intra-dataset testing}
\begin{table}[hbt]
\caption{Intra-dataset test results on the PURE, UBFC-rPPG and MMPD datasets. The unit is beats per minute (bpm). Best results are marked in bold and second best in underline.}
\label{tab:intra}
\centering
\resizebox{\textwidth}{!}{%
\begin{tabular}{l | c c c c | c c c c | c c c c}
\toprule
\textbf{Method} 
& \multicolumn{4}{c|}{\textbf{PURE}} 
  & \multicolumn{4}{c|}{\textbf{UBFC-rPPG}}
  & \multicolumn{4}{c}{\textbf{MMPD}} \\
             & MAE $\downarrow$ & MAPE $\downarrow$ & RMSE $\downarrow$ & $r$ $\uparrow$ & MAE $\downarrow$ & MAPE $\downarrow$ & RMSE $\downarrow$ & $r$ $\uparrow$ & MAE $\downarrow$  & MAPE $\downarrow$ & RMSE $\downarrow$ & $r$ $\uparrow$ \\
\midrule
LGI \cite{LGI}            & 3.59  & 3.37  & 14.66 & 0.79 & 5.39  & 11.08 & 15.09 & 0.81 & 16.63 & 23.06 & 18.77 & 0.11\\
CHROM \cite{CHROM}          & 5.39  & 11.08 & 15.09 & 0.81 & 4.06  & 3.34  & 8.83  & 0.89 & 13.73 & 18.88  & 16.95 & 0.15\\
POS \cite{POS}            & 0.36  & 0.5   & 0.93  & 1    & 4.08  & 3.93  & 7.72  & 0.92 & 15.61 & 21.40  & 18.28 & 0.14\\

PhysFormer \cite{PhysFormer}     & 0.52  & 0.86  & 1.03  & 0.99 & 2.34  & 2.60  & 5.55  & 0.97 & 13.64 & 19.39 & 14.42 & 0.15 \\

TS-CAN \cite{TS-CAN}     & 0.32  & 0.5  & 0.63  & 0.99 & 1.24  & 1.35  & 2.79  & 0.96 &  8.97 & 9.43 & 16.58 & 0.44 \\

EfficientPhys \cite{EfficientPhys}     & 0.55  & 0.71  & 1.34  & 0.99 & 0.73  & 0.83  & 2.53  & 0.97 & 12.79 & 13.48 & 21.12 & 0.24 \\

PhysMamba \cite{PhysMamba}      & 0.27  & 0.27 & 0.54  & 1 & 0.54  &  \underline{0.54} & \underline{0.79}  & 0.99 & - & - & - & -\\

RhythmFormer \cite{RhythmFormer}    & 0.27  & 0.31 & 0.46  & 1 & 0.81  & 0.80  & 1.12  & 0.99 & 6.74 & 6.93 & 11.92 & \underline{0.71}\\

LSTS \cite{LSTS} & \underline{0.15}  & \underline{0.22}  & \underline{0.40}  & 0.99    & \underline{0.51}     & 0.55   & 1.27     & 0.99    & \underline{4.80} & \underline{5.80} & \underline{10.46} & 0.69 \\
\midrule
\textbf{Reperio-rPPG (Ours)} & \textbf{0.07} & \textbf{0.12} & \textbf{0.21} & \textbf{1} & \textbf{0.15}  & \textbf{0.13}  & \textbf{0.51}  & \textbf{1} & \textbf{4.74} & \textbf{5.71}   & \textbf{10.16} & \textbf{0.72}\\
\bottomrule
\end{tabular}
} 
\end{table}
\paragraph{Evaluation on PURE dataset} We conducted experiments on the PURE \cite{PURE} dataset, which includes various conditions that mimic real-world scenarios. The models were evaluated using a subject-exclusive 5-fold cross-validation strategy \cite{LSTS}, where in each round, four folds were used for training and the remaining fold was used for testing. Notably, one subject in the PURE dataset exhibited exceptionally high pulse rates, so we excluded the round in which this subject was assigned to the test set, followed by \cite{LSTS}. The final results are reported as the average across all folds.
Table \ref{tab:intra} presents a comparative analysis between different methods. Traditional approaches such as LGI \cite{LGI} and CHROM \cite{CHROM} achieved suboptimal performance due to their reliance on restrictive feature extraction assumptions. In contrast, recent models based on Transformer architectures and State-Space Models demonstrate substantially improved results. Notably, methods such as TS-CAN \cite{TS-CAN}, PhysFormer \cite{PhysFormer}, RhythmFormer \cite{RhythmFormer}, and PhysMamba \cite{PhysMamba} consistently achieve strong metrics, with MAE, MAPE, and RMSE values all below 1.0. Surpassing these state-of-the-art methods, our proposed Reperio-rPPG establishes a new benchmark in the field. It not only achieves a substantial reduction in error—lowering MAE by 53\% (from 0.15 to 0.07) and RMSE by 48\% (from 0.40 to 0.21)—but also reaches a perfect Pearson correlation coefficient of 1.00, highlighting its exceptional accuracy, consistency, and robustness under varying facial expressions and motion artifacts.

\paragraph{Evaluation on UBFC-rPPG dataset}
Following previous studies \cite{PhysFormer, PhysMamba, RhythmFormer}, we divided the UBFC-rPPG dataset into a training set comprising the first 30 subjects and a test set comprising the remaining 12 subjects. As presented in Table \ref{tab:intra}, our proposed method achieves highly competitive performance on this dataset. To the best of our knowledge, no prior approach has attained such results. This outcome is primarily influenced by the dataset's inherent limitations, including its small size, limited diversity, and signal instability. Due to the restricted number of sample, existing rPPG models can easily fit the dataset, leading to a higher risk of overfitting. However, as we assess our model on more diverse and complex datasets, its true performance and generalization capabilities will become more apparent.
\paragraph{Evaluation on MMPD dataset}
Reperio-rPPG demonstrates exceptionally strong performance on PURE and UBFC-rPPG, confirming its effectiveness in controlled settings. However, real-world applications require robustness across diverse conditions. To further evaluate its generalization, we test on MMPD, a large-scale dataset featuring significant variations in skin tone, lighting, and activity. Following previous studies, we adopt a subject‐exclusive evaluation scheme, training on 26 subjects and testing on 7 unseen individuals. As summarized in Table \ref{tab:intra}, existing methods suffer substantial performance degradation due to the dataset’s complexity. In contrast, Reperio‐rPPG maintains superior robustness, achieving a MAE of 4.74, a RMSE of 10.16, and a Pearson correlation coefficient of 0.72. By addressing key challenges in rPPG measurement, this work brings contactless HR estimation one step closer to everyday use.
\paragraph{HRV Measurement}
\begin{figure}[H] 
    \centering
    \includegraphics[width=\linewidth]{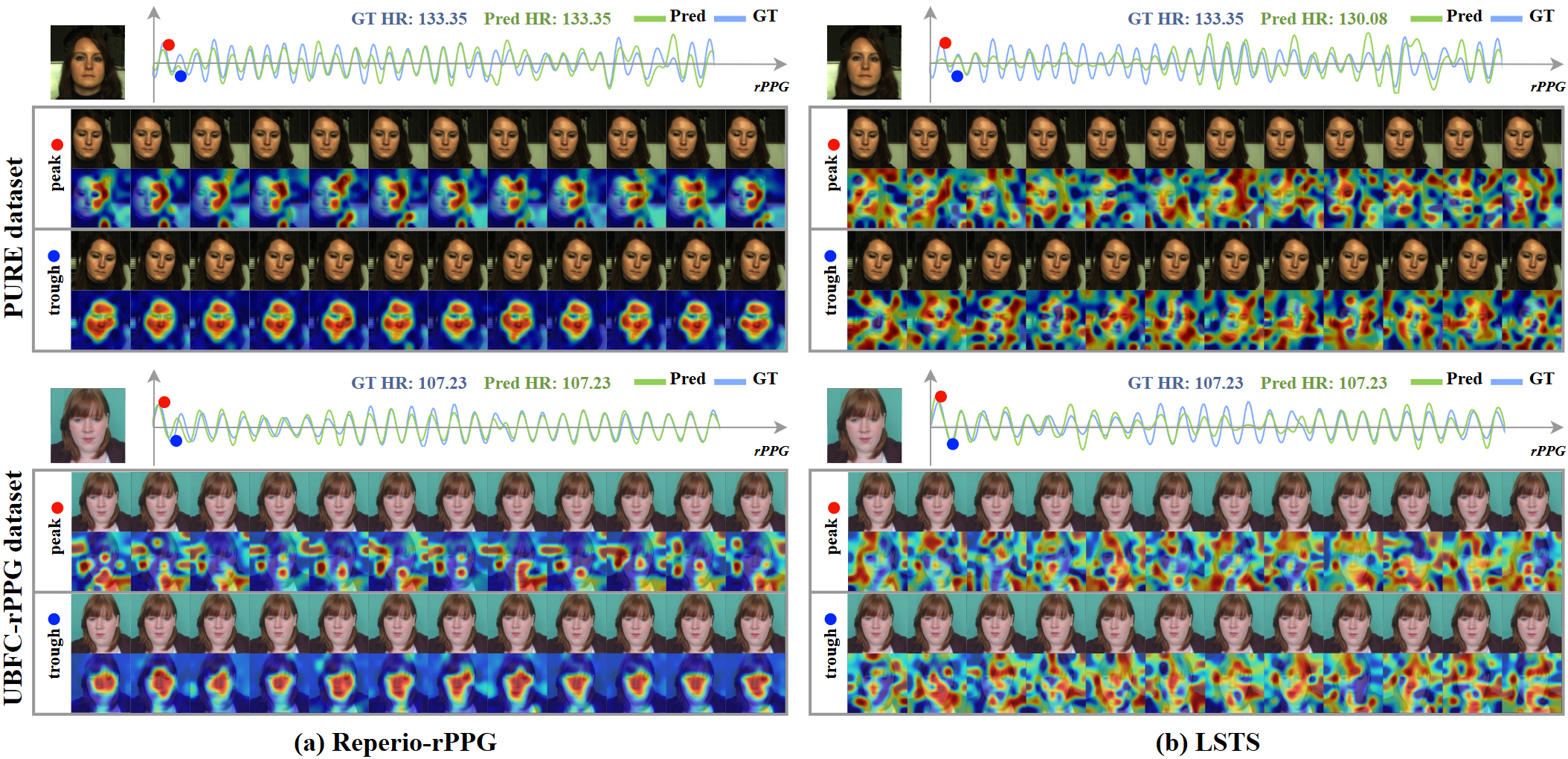} 
    \caption{The visualization illustrates the periodic characteristics of rPPG signals in two representative examples from the PURE and UBFC-rPPG datasets. The feature maps reveal a consistent pattern, with weak visual responses at the peaks of the rPPG signals and strong responses at the troughs. Specifically, peaks align with moments of maximum blood volume, resulting in the lowest reflected light intensity; conversely, troughs correspond to minimum blood volume, yielding the highest reflected intensity \cite{RhythmFormer} This inverse relationship underscores the periodic pulsations associated with blood flow dynamics. In comparing the two approaches, (a) Reperio-rPPG decouples spatial modeling from temporal modeling, thereby allowing its Transformer architecture to more effectively capture and concentrate on regions rich in HR information, such as key vascular areas on the face. By contrast, (b) LSTS incorporates a temporal shifting mechanism within the Transformer, which results in a broader dispersion of attention across the input, often extending to less relevant elements like hair, eyebrows, apparel, or the surrounding background.
    }
    \label{fig:CompareAttention} 
\end{figure}
\begin{table*}[hbt]
\centering
\caption{Comparison of HRV Measurement Performance on UBFC-rPPG and PURE Datasets.}
\label{tab:hrv_comparison}
\resizebox{\textwidth}{!}{%
\begin{tabular}{lcccccccccccc}
\toprule
 & \multicolumn{6}{c}{\textbf{UBFC-rPPG}} & \multicolumn{6}{c}{\textbf{PURE}} \\
\cmidrule(lr){2-7} \cmidrule(lr){8-13}
\multirow{2}{*}{\textbf{Method}} 
 & \multicolumn{2}{c}{\textbf{IBI (ms)}} & \multicolumn{2}{c}{\textbf{SDNN (ms)}} & \multicolumn{2}{c}{\textbf{SD1/SD2}} & \multicolumn{2}{c}{\textbf{IBI (ms)}} & \multicolumn{2}{c}{\textbf{SDNN (ms)}} & \multicolumn{2}{c}{\textbf{SD1/SD2}} \\
\cmidrule(lr){2-3} \cmidrule(lr){4-5} \cmidrule(lr){6-7} \cmidrule(lr){8-9} \cmidrule(lr){10-11} \cmidrule(lr){12-13} &
MAPE$\downarrow$ & $r\uparrow$ & MAPE$\downarrow$ & $r\uparrow$ & MAPE$\downarrow$ & $r\uparrow$ & MAPE$\downarrow$ & $r\uparrow$ & MAPE$\downarrow$ & $r\uparrow$ & MAPE$\downarrow$ & $r\uparrow$ \\
\midrule
TS-CAN \cite{TS-CAN} & 3.05 & 0.87 & 107.15 & 0.26 & 81.17 & 0.40 & 2.51 & 0.91 & 97.53 & 0.29 & 78.41 & 0.42 \\
EfficientPhys \cite{EfficientPhys} & 4.12 & 0.85 & 132.19 & 0.15 & 87.52 & 0.21 & 3.71 & 0.90 & 127.75 & 0.16 & 86.46 & 0.21 \\
PhysFormer \cite{PhysFormer} & 0.85 & 0.99 & 22.10 & 0.61 & 26.66 & 0.71 & 0.61 & 1.00 & \textbf{16.27} & \textbf{0.91} & 24.15 & 0.73 \\
PhysNet \cite{PhysNet} & 1.23 & 0.99 & 25.91 & 0.59 & 40.37 & 0.61 & 0.67 & 0.99 & 30.77 & 0.72 & 37.43 & 0.63 \\
LSTS \cite{LSTS} & \underline{0.53} & \underline{1.00} & \underline{17.03} & \underline{0.66} & \textbf{10.93} & \textbf{0.90} & \underline{0.52} & \underline{1.00} & 19.22 & 0.80 & \underline{20.79} & \underline{0.78} \\
\textbf{Reperio-rPPG (Ours)}  & \textbf{0.43} & \textbf{1.00} & \textbf{15.60} & \textbf{0.66} & \underline{12.97} & \underline{0.89} & \textbf{0.41} & \textbf{1.00} & \underline{18.29} & \underline{0.83} & \textbf{18.78} & \textbf{0.81} \\
\bottomrule
\end{tabular}
}
\end{table*}
The experimental results for HRV measurement are illustrated in Table~ \ref{tab:hrv_comparison}. It is shown that our Reperio-rPPG achieves more accurate HRV measurement on 8 out of the 12 metrics across the UBFC-rPPG and PURE datasets. HRV reflects the variations in the time interval between consecutive heartbeats. Therefore, the HRV measurement results demonstrate that our proposed methods can effectively capture the inter-period variations in physiological signals. Notably, Reperio-rPPG outperforms previous methods on the nonlinear SD1/SD2 measurement on the PURE dataset by a significant margin, and achieves competitive performance on the UBFC-rPPG dataset. Since SD1/SD2 reflects the frequency-domain characteristics of HRV, this result suggests that our method is more effective than previous studies in uncovering the periodicity of physiological signals.
\subsubsection{Cross-dataset testing}
\begin{figure}[hbt] 
    \centering
    \includegraphics[width=0.7\linewidth]{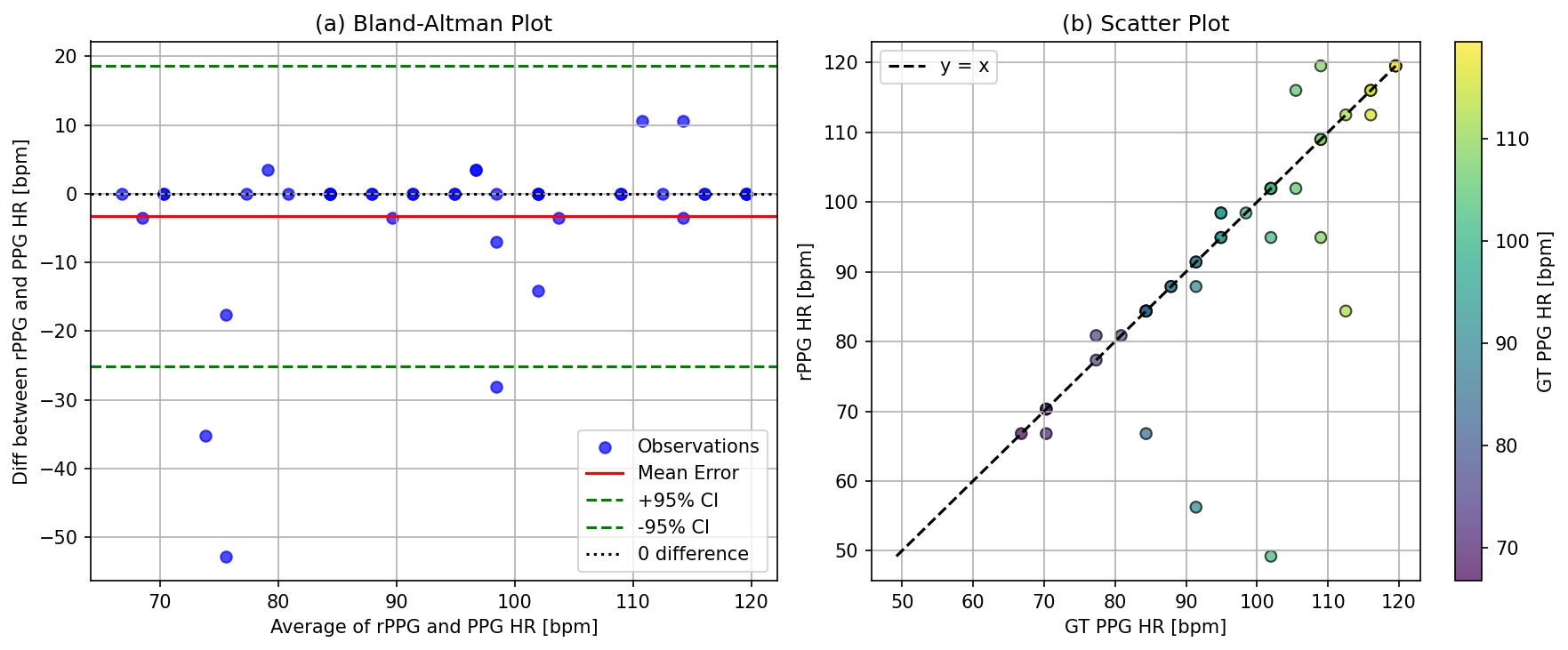} 
    \caption{The Bland-Altman plot (a) and Scatter plot (b) show the difference between estimated HR and GT HR on the cross-dataset evaluation (PURE~$\rightarrow$~UBFC-rPPG).}
    \label{fig:scatter_bland_altman_hr_pure_ubfc} 
\end{figure}
\begin{figure}[hbt] 
    \centering
    \includegraphics[width=0.7\linewidth]{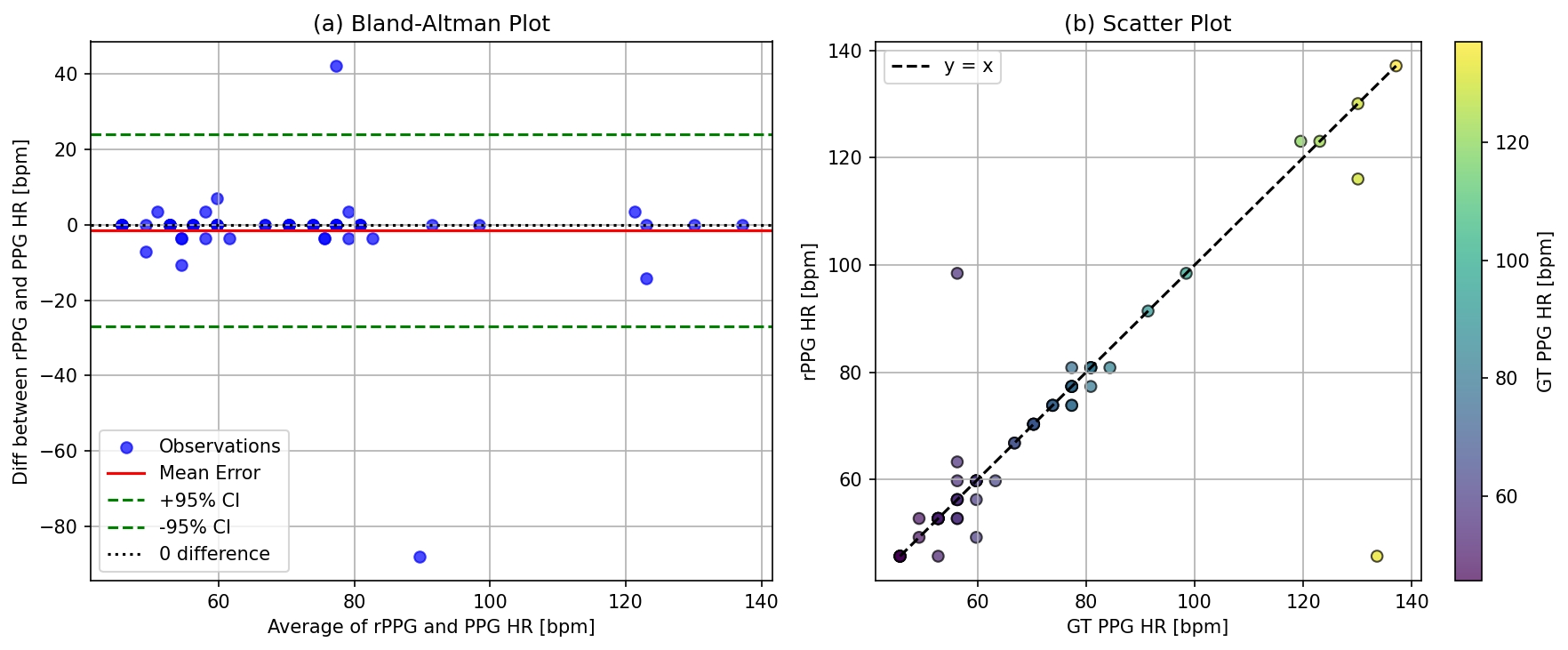} 
    \caption{The Bland-Altman plot (a) and Scatter plot (b) show the difference between estimated HR and GT HR on the cross-dataset evaluation (UBFC-rPPG~$\rightarrow$~PURE).}
    \label{fig:scatter_bland_altman_hr_ubfc_pure} 
\end{figure}
\begin{table}[hbt]
\caption{Cross-dataset test results. MMPD$\rightarrow$UBFC-rPPG means the metric trained on MMPD and tested on UBFC-rPPG.}
\label{tab:cross-comparison}
\centering
\resizebox{\textwidth}{!}{%
\begin{tabular}{l|ccc|ccc|ccc|ccc}
        \midrule
        \multirow{2}{*}{\textbf{Method}} & \multicolumn{3}{c|}{\textbf{MMPD$\rightarrow$UBFC-rPPG}} & \multicolumn{3}{c|}{\textbf{MMPD$\rightarrow$PURE}} & \multicolumn{3}{c|}{\textbf{PURE$\rightarrow$UBFC-rPPG}} & \multicolumn{3}{c}{\textbf{UBFC-rPPG$\rightarrow$PURE}} \\
        
        & MAE$\downarrow$ & RMSE$\downarrow$ & $r \uparrow$ & MAE$\downarrow$ & RMSE$\downarrow$ & $r \uparrow$ & MAE$\downarrow$ & RMSE$\downarrow$ & $r \uparrow$ & MAE$\downarrow$ & RMSE$\downarrow$ & $r \uparrow$ \\
        \midrule
        EfficientPhys \cite{EfficientPhys} & 6.55  & 13.95 & 0.72  & \underline{6.20} & \underline{15.70}  & \underline{0.74}  & 8.30  & 21.05 & 0.55 & 7.20 & 17.86 & 0.69\\
        PhysMamba \cite{PhysMamba} & -  & - & -  & - & -  & -  & 2.07  & 3.46 & \underline{0.99} & 1.24 & 5.02 & 0.98\\
        RhythmFormer \cite{RhythmFormer} & 3.23  & 5.71 & 0.94  & 12.76 & 22.45  & 0.40  & 1.60  & \underline{3.37} & 0.98 & 3.14 & 9.65 & 0.92\\
        TS-CAN \cite{TS-CAN} & -  & - & - & 7.24 & 16.80  & 0.68  & 6.94  & 17.14 & 0.67 & 7.62 & 20.46 & 0.62\\
        LSTS \cite{LSTS} & \underline{1.03} & \underline{2.50} & \underline{0.99} & 8.91 & 21.12 & 0.46
         & \underline{1.51}  & 3.39 & 0.98 & \textbf{0.25} & \textbf{0.69} & \underline{0.99}\\

\midrule
        \textbf{Reperio-rPPG (Ours)}  & \textbf{1.03} & \textbf{2.44} & \textbf{0.99} & \textbf{2.29} & \textbf{7.65} & \textbf{0.95} & \textbf{1.15} & \textbf{2.48} & \textbf{0.99} & \underline{0.50} & \underline{1.20} & \textbf{1} \\
\bottomrule
\end{tabular}
} 
\end{table}
To further evaluate the generalization capability of our method, we conducted cross-dataset testing using the PURE, UBFC-rPPG, and MMPD datasets. In Table \ref{tab:cross-comparison}, we compared the performance of our model against state-of-the-art methods: EfficientPhys~\cite{EfficientPhys}, PhysMamba~\cite{PhysMamba}, RhythmFormer~\cite{RhythmFormer}, TS-CAN~\cite{TS-CAN} and LSTS~\cite{LSTS}.

The experiments were divided into four settings: MMPD → PURE, MMPD → UBFC-rPPG, PURE → UBFC-rPPG, and UBFC-rPPG → PURE. In each case, the model was trained on one dataset (source domain) and tested on another (target domain). Among these datasets, MMPD provides the most diverse recording conditions, covering a wider range of real-world variations compared to PURE and UBFC-rPPG. As a result, training on MMPD generally leads to better model performance on other datasets, since the model learns from more complex scenarios. Fig. \ref{fig:scatter_bland_altman_hr_pure_ubfc} and \ref{fig:scatter_bland_altman_hr_ubfc_pure} present, respectively, the Bland–Altman and scatter plots for the PURE → UBFC-rPPG and UBFC-rPPG → PURE transfer settings. In the Bland–Altman plots, the mean bias remains close to zero, suggesting no significant systematic difference between estimated and reference HR. Meanwhile, the scatter plots show that the predicted rPPG HR values cluster tightly along the identity line, confirming strong linear correlation and high precision even under unseen recording conditions. These visualizations and performance from Table \ref{tab:cross-comparison}  
underscore the robustness and adaptability of our model when applied to new domains without any domain-specific fine-tuning.
\section{Ablation Study}

\begin{table}[hbt]
\caption{Ablation Study of Data Augmentation Techniques on the
MMPD Dataset}
\label{tab:ablation_augmentations}
\centering
\setlength{\tabcolsep}{8pt}
\renewcommand{\arraystretch}{1.2}
\scalebox{0.8}{
\begin{tabular}{c|c|c| c c c c}
\toprule
TCM & NDF & MPOS & MAE$\downarrow$ & MAPE$\downarrow$ & RMSE$\downarrow$ & $r \uparrow$ \\
\midrule
 &  &   & 9.12 & 10.58 & 16.70 & 0.32 \\
 &  & \ding{51}  & 7.23 & 8.23 & 14.19 & 0.44 \\
 & \ding{51} &   & 7.24 & 8.47 & 15.33 & 0.43 \\
 & \ding{51} & \ding{51} & 6.64 & 7.38 & 13.58 & 0.54 \\
\ding{51} &  &  & 6.47 & 7.45 & 13.12 & 0.56 \\
\ding{51} &  & \ding{51} & 6.19 & 7.07 & 12.69 & 0.58 \\
\ding{51}  & \ding{51}  &  & 5.63 & 6.48 & 10.98 & 0.67 \\
\ding{51}  & \ding{51}  & \ding{51} & \textbf{4.74} & \textbf{5.71} & \textbf{10.16} & \textbf{0.72} \\
\bottomrule
\end{tabular}
}
\end{table}
This section presents ablation studies to systematically assess the contribution of each proposed module within the Reperio-rPPG framework.
\subsection{Impact of CutMix}
\begin{table}[hbt]
\caption{Ablation Study on UBFC-rPPG and PURE Datasets with TCM Augmentation}
\label{tab:ablation_cutmix}
\centering
\resizebox{\textwidth}{!}{%
\begin{tabular}{l | c | c c c c | c c c c}
\toprule
\textbf{Method} & \textbf{CutMix} & \multicolumn{4}{c|}{\textbf{UBFC-rPPG}} & \multicolumn{4}{c}{\textbf{PURE}} \\
                &                 & MAE $\downarrow$ & MAPE $\downarrow$ & RMSE $\downarrow$ & $r$ $\uparrow$
                                  & MAE $\downarrow$ & MAPE $\downarrow$ & RMSE $\downarrow$ & $r$ $\uparrow$ \\
\midrule
LSTS & \ding{55} & 0.51 & 0.55 & 1.27 & 0.99  & 0.15 & 0.22 & 0.40 & 0.99 \\
LSTS & \ding{51} & 0.22 & 0.25 & 0.76 & 1 & 0.14 & 0.21 & 0.40 & 0.99 \\
Reperio-rPPG (Ours) & \ding{55} & 0.15 & 0.14 & 0.51 & 1 & 0.09 & 0.14 & 0.24 & 1 \\
Reperio-rPPG (Ours) & \ding{51} & \textbf{0.15} & \textbf{0.13} & \textbf{0.51} & \textbf{1} & \textbf{0.07} & \textbf{0.12} & \textbf{0.21} & \textbf{1} \\
\bottomrule
\end{tabular}
}
\end{table}

A major limitation of existing rPPG datasets is their lack of diversity, which often leads to overfitting and limited generalization capability. With limited samples and controlled recording conditions, models tend to learn dataset-specific patterns rather than robust features. To address this issue, we adapted the CutMix augmentation technique, originally designed for image data, and applied it to video-based rPPG estimation. As shown in Table \ref{tab:ablation_cutmix}, integrating CutMix significantly enhances the performance of Reperio-rPPG, especially on small-scale datasets such as UBFC-rPPG \cite{UBFC} and PURE\cite{PURE}. This augmentation also proves effective on more complex datasets like MMPD~\cite{MMPD}, as detailed in Table~\ref{tab:ablation_augmentations}, where combining CutMix with NDF and MPOS achieves state-of-the-art results. Moreover, when we applied CutMix to LSTS \cite{LSTS}, we observed consistent performance gains, demonstrating that the benefits of CutMix extend beyond our own approach.

\subsection{Impact of Swin Transformer Backbone}
\begin{figure}[hbt] 
    \centering
    \includegraphics[width=0.6\linewidth]{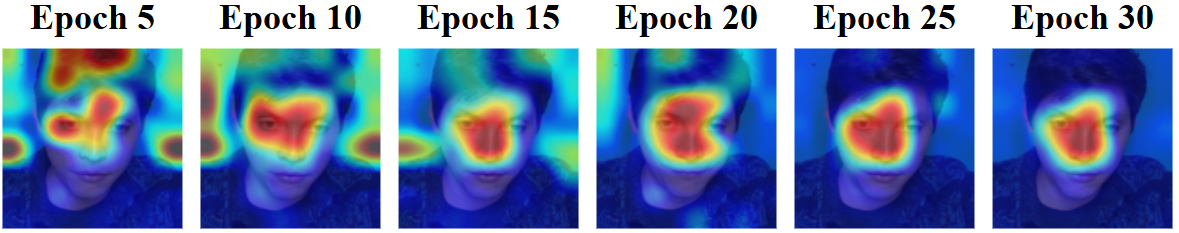} 
    \caption{Attention convergence across epochs in the final Swin Transformer block, evaluated on UBFC-rPPG dataset.}
    \label{fig:ConvergenceAttention} 
\end{figure}
In this section, we analyze the role of the Swin Transformer in spatial representation learning using the UBFC-rPPG dataset \cite{UBFC}. To examine how spatial attention evolves during training, we visualize attention maps from the final Swin Transformer block at different epochs, as shown in Fig.~\ref{fig:ConvergenceAttention}. The model gradually refines its attention from broad, diffuse regions to more discriminative areas, eventually focusing on physiologically relevant facial regions associated to BVP signals, such as the forehead, cheeks, and nose. This evolution demonstrates a clear convergence toward meaningful spatial priors that facilitate accurate rPPG estimation.

The effectiveness of our spatial modeling strategy is further validated through a comparison with the state-of-the-art LSTS baseline\cite{LSTS}, which also employs the Swin Transformer but incorporates a temporal shift mechanism into the attention computation. As illustrated in Fig.~\ref{fig:CompareAttention}, our model exhibits more localized and precise attention on key facial regions, whereas LSTS ~\cite{LSTS} tends to distribute attention more broadly, including non-informative areas such as hair, eyebrows, clothing, or background. This comparison highlights the benefit of decoupling temporal modeling from the spatial attention mechanism, enabling Reperio-rPPG to suppress irrelevant features and focus more directly on regions containing high HR signal information.

\subsection{Impact of R-GCN and Graph Transformer}
\begin{table}[hbt]
\caption{Ablation Study on the UBFC-rPPG Dataset Using R-GCN and Graph Transformer Modules}
\label{tab:ablation_rgcn_graphtrans}
\centering
\renewcommand{\arraystretch}{1.1} 
\scalebox{0.8}{
\begin{tabular}{c | c | c c c c}
\toprule
\textbf{Graph Transformer} & \textbf{R-GCN} &  \textbf{MAE $\downarrow$} & \textbf{MAPE $\downarrow$} & \textbf{RMSE $\downarrow$} & $r$ $\uparrow$
                 \\
\midrule
\ding{51} & \ding{55} & 0.59 & 0.61 & 1.19 & 0.99 \\
\ding{55} & \ding{51} & 0.37 & 0.35 & 0.91 & 1 \\
\ding{51} & \ding{51} & \textbf{0.15} & \textbf{0.13} & \textbf{0.51} & \textbf{1} \\
\bottomrule
\end{tabular}
}
\end{table}
\begin{table}[hbt]
\caption{Ablation Study of Temporal Window Configurations on the MMPD Dataset}
\label{tab:ablation_temporal_windows}
\centering
\setlength{\tabcolsep}{9pt} 
\scalebox{0.9}{
\begin{tabular}{c|c|c|c c c c}
\toprule
$\mathcal{P}$ & $\mathcal{F}$ & $\Delta$ & MAE $\downarrow$ & MAPE $\downarrow$ & RMSE $\downarrow$ & $r \uparrow$ \\
\midrule
 &  & \ding{51}       & 8.21 & 9.66 & 15.42 & 0.43 \\
 & \ding{51} &        & 7.58   & 8.51   & 15.46  & 0.54   \\
 & \ding{51} & \ding{51} & 6.05 & 7.12 & 13.00  & 0.56 \\
\ding{51} &  &        & 6.67  & 7.69 & 12.57 & 0.63 \\
\ding{51} &  & \ding{51} & 5.51 & 6.41  & 10.7 & 0.72 \\
\ding{51} & \ding{51}  &  & 6.70 & 8.06  & 14.27 & 0.59 \\
\ding{51} & \ding{51} & \ding{51} & \textbf{4.74}   & \textbf{5.71}   & \textbf{10.16}    & \textbf{0.72}   \\
\bottomrule
\end{tabular}
}
\end{table}
We begin by evaluating the individual and joint contributions of the R-GCN and Graph Transformer modules on the UBFC-rPPG dataset (Table~\ref{tab:ablation_rgcn_graphtrans}). When applied independently, the Graph Transformer yields a relatively high MAE of 0.59 and RMSE of 1.19, despite achieving a strong correlation coefficient of 0.99. In contrast, R-GCN alone substantially improves performance, reducing the MAE to 0.37 and RMSE to 0.91, while attaining a perfect Pearson correlation of 1.00. This demonstrates the benefit of incorporating relational graph structures to more effectively capture temporal dependencies in physiological signals. The best results are obtained when both modules are integrated—the combined model achieves the lowest errors across all metrics (MAE of 0.15, MAPE of 0.13, and RMSE of 0.51), along with a perfect correlation coefficient of 1.00. These findings highlight the strong synergy between the message-passing mechanism of R-GCN and the attention-based refinement of the Graph Transformer, resulting in highly accurate and robust pulse estimation.

Turning to the MMPD dataset (Table~\ref{tab:ablation_temporal_windows}), we shift focus to the internal temporal design of the R-GCN. Here, we compare various windowing strategies and find that integrating past, future, and periodic windows yields the best performance, achieving the lowest MAE of 4.74. This configuration outperforms all single-window strategies by a substantial margin, with improvements of up to 42\%.

An important question is whether these gains incur a high computational cost. As summarized in Table~\ref{tab:complexity_mmpd}, despite integrating both R-GCN and a Graph Transformer, Reperio-rPPG remains lightweight, containing only 6.63M parameters and requiring 66.31 GMACs, comparable to or more efficient than strong baselines. Furthermore, it supports real-time inference on commodity hardware: on an NVIDIA T4 GPU, a single-sample forward pass completes in 93.22 ms, with peak GPU memory of 0.48 GB (PyTorch max-allocated).
\begin{table}[hbt]
\centering
\caption{Computational Complexity Analysis on MMPD Dataset}
\label{tab:complexity_mmpd}
\scalebox{0.8}{
\begin{tabular}{lccc}
\toprule
\textbf{Model} & \makecell{\textbf{Number of} \\ \textbf{Parameters (M)}} & \textbf{MACs (G)} & \textbf{MAE} \textbf{($\downarrow$)} \\
\midrule
TS-CAN \cite{TS-CAN}              & 2.23 & 40.62 & 8.97 \\
EfficientPhys \cite{EfficientPhys} & 2.16 & 20.39 & 12.79 \\
PhysFormer \cite{PhysFormer}       & 7.38 & 56.32 & 13.64 \\
LSTS \cite{LSTS}                  & 6.37 & 66.34 & 4.80 \\
\textbf{Reperio-rPPG (Ours)}           & 6.63 &  66.31 & 4.74 \\
\bottomrule
\end{tabular}
}
\end{table}

In summary, our results highlight that both the temporal structuring within R-GCN and its integration with attention-based transformers are critical for achieving state-of-the-art accuracy in remote photoplethysmography. The model exhibits strong generalization and precision across different conditions and dataset domains.

\section{Conclusion}
In this work, we introduce Reperio-rPPG, a novel network architecture for rPPG signal estimation. Our framework integrates two key components: the Swin Transformer for capturing robust, multi-level spatial information, and the combination of R-GCN and Graph Transformer for effectively modeling the intrinsic periodic patterns present in physiological signals. Through comprehensive evaluations conducted on three widely-used benchmark datasets—PURE \cite{PURE}, UBFC-rPPG \cite{UBFC}, and MMPD\cite{MMPD}—Reperio-rPPG demonstrates state-of-the-art performance, surpassing existing methods in remote physiological signal measurement. Moreover, we perform detailed ablation studies to assess the individual contributions of each architectural component, confirming the effectiveness and necessity of both spatial and temporal modeling modules. Despite the promising results, several limitations remain, including the need to predefine a fixed temporal window size. Furthermore, as visualized in Fig.~\ref{fig:heatmap}, the model can still be challenged by factors such as skin tone diversity, facial expression variations, and extreme motion. These observations suggest valuable directions for future research — including the incorporation of adaptive temporal modeling, improved motion compensation, and fairness-driven domain generalization strategies to enhance robustness across diverse real-world conditions and subject populations.
\bibliographystyle{elsarticle-num} 
\bibliography{bibliography}
\end{document}